\documentclass{article}

\usepackage{microtype}
\usepackage{graphicx}
\usepackage{subfigure}
\usepackage{booktabs} 
\usepackage{multirow}
\usepackage{paralist}
\usepackage{todonotes}
\usepackage{amsmath}
\usepackage{amsfonts}
\usepackage{bbm}
\usepackage{gensymb}
\usepackage{capt-of}
\usepackage[draft]{hyperref}

\usepackage[accepted]{icml2019}
\def \cX {\mathcal{X}}
\def \cY {\mathcal{Y}}
\def \cD {\mathcal{D}}

\icmltitlerunning{Low-Dimensional Visual Attribute (LDVA) Encoders}

\begin{document}

\twocolumn[
\icmltitle{Learning Classifiers for Domain Adaptation, Zero and Few-Shot Recognition Based on Learning Latent Semantic Parts}



\icmlsetsymbol{equal}{*}

\begin{icmlauthorlist}
\icmlauthor{Pengkai Zhu}{equal,bu}
\icmlauthor{Hanxiao Wang}{equal,bu}
\icmlauthor{Venkatesh Saligrama}{bu}
\end{icmlauthorlist}

\icmlaffiliation{bu}{Electrical and Computer Engineering Department, Boston University, Boston, USA}


\icmlkeywords{Machine Learning, ICML}

\vskip 0.3in
]



\printAffiliationsAndNotice{\icmlEqualContribution} 

\begin{abstract}
In computer vision applications, such as domain adaptation (DA), few shot learning (FSL) and zero-shot learning (ZSL), we encounter new objects and environments, for which insufficient examples exist to allow for training ``models from scratch,'' and methods that adapt existing models, trained on the presented training environment, to the new scenario are required. We propose a novel visual attribute encoding method that encodes each image as a low-dimensional probability vector composed of prototypical part-type probabilities. The prototypes are learnt to be representative of all training data. 
At test-time we utilize this encoding as an input to a classifier. At test-time we freeze the encoder and only learn/adapt the classifier component to limited annotated labels in FSL; new semantic attributes in ZSL. We conduct extensive experiments on benchmark datasets. Our method outperforms state-of-art methods trained for the specific contexts (ZSL, FSL, DA).    
\end{abstract}

\vspace{-0.5cm}
\section{Introduction}
\label{intro}
Deep Neural Networks have emerged as the state-of-art in terms of achievable accuracy in a wide variety of applications including large-scale visual classification. Nevertheless, this success of DNNs has critically hinged on availability of large amount of labeled training data, data that is labeled by human labelers. It is increasingly being recognized, particularly in the context of large-scale visual classification problems~\cite{ILSVRCarxiv14}, that such large-scale human labeling is not scalable \cite{antol2014zero}, and we must account for challenges posed by non-uniform and sparsely annotated training data \cite{Bhatia15} as in few-shot learning (FSL), appearance of novel objects for in-the-wild scenarios as in generalized zero-shot learning (GZSL), and responding to changes in operational environment such as changes in data collection viewpoints as exemplified by domain adaptation (DA).  

\textit{Decomposability and Compositionality:} We are motivated to respond to these aforementioned challenges without training ``models from scratch,'' which requires collecting new labeled data, and yet achieving high-accuracy. We propose a novel framework and DNN architecture that addresses these challenges in a unified manner. Our key insight is based on decomposability of objects into proto-typical primitive parts/part-types and compositionality of proto-typical primitive part/part-types to explain new, unseen or modified object classes. This insight is not new and has been employed in a long-line of work particularly in cognitive science to explain human concept learning such as children making meaningful generalization through one-shot learning, parsing objects into parts, and generating new concepts from parts (see ~\cite{lake2015human}). While \cite{lake2015human} advocates a generative Bayesian Program Learning framework to mimic human concept learning and avoid ``data-hungry'' DNNs altogether, we advocate use of DNNs and situate our work within a discriminative learning framework and employ novel DNN architectures that also obviates the need for new annotated data and realizes high-accuracy.

\textit{Our Contributions.}
We propose a novel approach that encodes an input instance as a collection of probability vectors. Each probability vector is associated with a part and represents the mixture of prototypical part types that makeup the part. To do this we train a Multi-Attention CNN (MACNN), which produces a diverse collection of attention regions and associated features masking out uninteresting regions of the image space. These attention regions are decomposed into a suitably small number of prototypical parts and prototypical part probabilities, yielding a low-dimensional encoding. We refer to these encodings as low-dimensional visual attribute (LDVA) encodings since they are analogous to how humans would quantify the existence of an attribute in the presented instance by drawing similarity to a prototypical attribute seen from experience. We input the LDVA encoding into a predictor, which then predicts the output for the different scenarios (ZSL, FSL or DA). We learn an end-to-end model on training data and at test-time, freeze the high-dimensional mapping to LDVA encoding component, and only adapting the predictor based on what is revealed during test-time. 

\textit{Training \& Test-time Prediction:} For unsupervised domain adaptation (UDA), both annotated source data and unannotated target data are utilized for training our end-to-end model. However, since no additional data is available, both LDVA and the classifier are unchanged at test-time. For GZSL, we assume a one-to-one correspondence between semantic vectors and class labels as is the convention. During training we assume access to seen class image instances and associated semantic vectors, while being agnostic to both unseen images and unseen semantic vectors. At test-time, we fix our LDVA embedding and modify the prediction component to incorporate semantic vectors from all seen and unseen classes. Finally, for FSL, we learn only the classifier using LDVA as inputs. 

\textit{Why is LDVA effective?} Our results on benchmark datasets highlights the utility of mapping visual instances into the LDVA
encoding and its tolerance to visual distortions. For DA, while an image can exhibit significant visual distortion and so domain shifts, the LDVA encodings for source and target are similar\footnote{consider handwritten digits under going a domain shift but the composition of parts that make up the digit is quite similar}, thus obviating the need to modify the classifier (see {\it Figure~\ref{fig:pi}}). For GZSL, the LDVA encoding mirrors how semantic attributes are scored. This enables meaningful knowledge transfer from visual to semantic domain and reducing the semantic gap (see {\it Figure~\ref{fig:embedding}}).

Section 2 describes related work. In Section 3 we first present an overview of proposed approach and then later describe concretely our models. In Section 4 we describe experiments on benchmark datasets for DA, FSL and ZSL. 

\vspace{-0.2cm}
\section{Related Work}
\label{related}
Related approaches for adapting models from presented training environment (PTE) can be divided into three groups: pixel-space methods, feature-space methods, and latent-space methods. In contrast to our LDVA encoding that encodes the mixture composition of parts, these works typically attempt to transfer knowledge in a high-dimensional space. We list different lines of research in this context. 

{\it Pixel-space methods} focus on generating or synthesizing images in the new visual environments, or converting new environment images into existing PTE, so as to avoid exhaustive human annotation. These works are largely based on the recently proposed Generative Adversarial Networks~\cite{goodfellow2014generative}.
In domain adaptation, \cite{taigman2016unsupervised,shrivastava2017learning,bousmalis2017unsupervised}
propose to train a generator to transform a source image into a target image (or vice versa) and meanwhile force the generated image to be similar to the original one.
\cite{liu2016coupled} trains a tuple of GANs for both domains and ties the weights for certain layers to jointly learn a source and target representation.
~\cite{ghifary2016deep} enforces the features learnt on the source data to reconstruct target images to encourage alignment in the unsupervised adaptation setting.
In generalized zero-shot learning, analogous to domain adaptation, attempts have been made on synthesizing unseen class images in the new environment from the given semantic attributes, e.g. \cite{zhu2018generative,Verma_2018_CVPR,Xian_2018_CVPR,Jiang_2018_ECCV}.
In few-shot learning, generative models are often used for data augmentation to account for the sparsely labelled few-shot examples, e.g.
\cite{antoniou2017data,wang2018low,mehrotra2017generative}.

{\it Feature-space methods} propose to either directly learn environment-invariant feature/predictor model, or align the models from the new and source environments to address the problem of insufficient annotations.
For instance, several domain adaptation works propose learning a domain-invariant feature embedding via adversarial training~
\cite{long2018conditional,tzeng2017adversarial} and graph-based label propagation~\cite{ding2018graph}, while others 
propose aligning the target domain feature distribution to source domain, e.g. \cite{kumar2018co}.
There are also methods which perform adaptation in both feature-space and pixel-space. 
For example, \cite{hoffman2017cycada} proposes a model which adapts between domains using both generative image space alignment and latent representation space alignment.
Similar approaches have also been investigated in GZSL. 
\cite{frome2013devise,Lee_2018_CVPR,Wang_2018_CVPR} propose learning feature embeddings that directly map the visual domain to the semantic domain
and infer classifiers for unseen classes.
In \cite{Annadani_2018_CVPR, kodirov2017semantic}, authors propose an encoder-decoder network with the goal of mirroring learnt semantic relations between different classes in the visual domain.
In FSL,  \cite{sung2018learning} and \cite{vinyals2016matching} propose adopting an environment-invariant feature representation that is based on comparing an input sample to a support set and use the similarity scores for classification input in the new environment .

{\it Latent-space methods} aim to discover latent feature spaces for PTE images that are universal and agnostic to environment changes, and thus can be further used as a general representation for newly encountered images in a new environment. For example, these latent variables include locations of the attention regions on interesting foreground objects, clusters and manifolds information of the data distributions, or common visual part features (which are still high-dimensional).
For DA, \cite{kang2018deep} assumes attention of the convolutional layers to be invariant to the domain shift and propose aligning the attentions for source and target domain images. 
\cite{wang2018visual} learns Grassmann manifold with structural risk minimization, and train a domain-invariant classifier on the learnt manifold.
\cite{shu2018dirt} makes use of the data distribution by first clustering the data and assumes samples in the same cluster share the same label. Target domain is modified so as to not break clusters.
In GZSL, \cite{Li_2018_CVPR} propose zoom-net as a means to filter-out redundant visual features such as deleting background and focus attention on important locations of an object. \cite{zhu2018generative} further extend this insight and propose visual part detector (VPDE-Net) and utilize high-dimensional part feature vectors as an input for semantic transfer, namely, to synthesize unseen examples by leveraging knowledge of unseen class attributes. Similarly in FSL, \cite{snell2017prototypical} propose learning prototypical representations of each class by  hard clustering on a support set and perform classification on these representations. 
\cite{lin2017transfer} finds that training manifolds in 3D views results in manifolds that are more general and abstract, likely at the levels of parts, and independent of the specific objects or categories in the PTE. There are also the family of meta-learning methods, e.g. \cite{ravi2016optimization, munkhdalai2017meta, finn2017model} which treats the model/optimization parameters as latent variables and propose meta-models to infer such parameters.

\vspace{-0.2cm}
\section{Methodology}
\label{method}

\begin{figure*}
\centering
\includegraphics[width=0.95\textwidth]{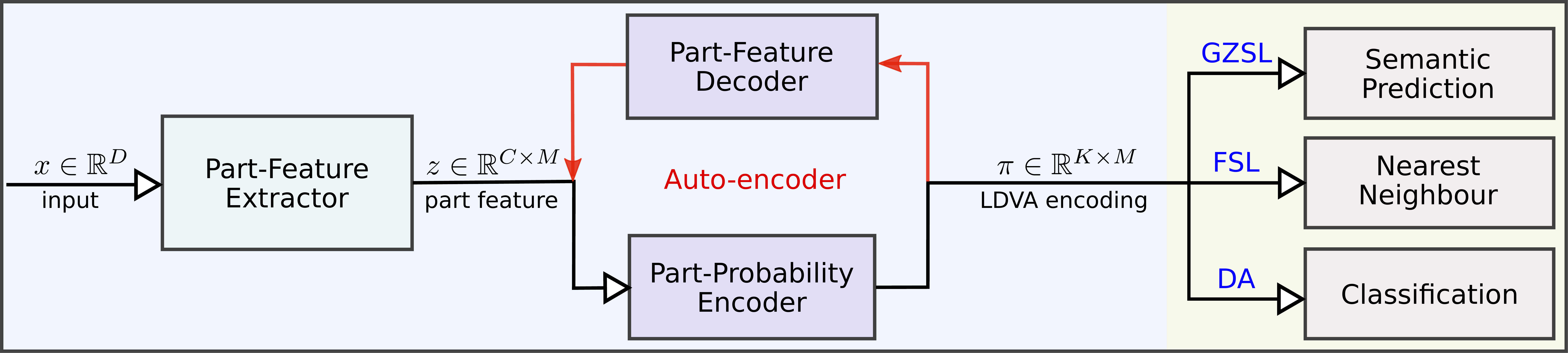}
\caption{The proposed network architecture. For an input image $x$, the part feature extractor decompose $x$ into $M$ parts and extracts associated features $z_m$, the part-probability encoder then encodes each part feature as a low-dimensional encoding $\pi_m$ by projecting the features onto a dictionary of primitive proto-typical part-types automatically discovered by our model. $\pi_m$ is then used as inputs to train task-specific predictor models for GZSL, FSL and DA.}
\label{fig:model}
\end{figure*}

\subsection{Problem Definition}
The problem scenarios under consideration consist of source and target domains and call for prediction on target domain by means of training data available in different forms. 
We denote by $x \in \cX \subset \mathbb{R}^D$ inputs taking values in a feature space and $y \in \cY$ the output labels taking values in a finite set $\cY$ and $p(x,y)$ the joint distribution. Whenever necessary, we denote by $p_s(x,y),\,p_t(x,y)$ source and target joint distributions respectively. Since we focus primarily on images of fixed dimension, we assume that the input space for source and target domain is the same. We allow for class labels for source and target domains to be different. We use superscript notation: $y^s, \cY^s, y^t, \cY^t$ when necessary for source and target labels to avoid confusion. 

\noindent
\textbf{Unsupervised Domain Adaptation (UDA)}: Source and target domain spaces share same labels. The joint distributions $p_s(y|x) \neq p_t(y|x)$ and $p_s(x|y) \neq p_t(x|y)$. For training, we are provided $n_s$ IID instances of annotated source domain data $(x_i,y_i)\stackrel{d}{\sim} p_s(x,y),\,\,i\in[n_s]$ and $n_t$ IID instances of unannotated input instances $x_i \stackrel{d}{\sim} p_t(x)=\sum_{y \in \cY} p_t(x,y)$. Our goal is to learn a predictor $f(\cdot)$ that generalizes well, i.e., the expected loss $\bar{L}_t=\mathbb{E}_{(x,y)\sim p_t} \mathbbm{1}_{\{f(x)\neq y\}}$ is small. 

\noindent
\textbf{Few Shot Learning (FSL)}: 
Note that while UDA and FSL share some similarities, i.e., $p_{s}(x,y) \neq p_t(x,y)$, they are different cases because, in FSL, the collection of source and target labels are not identical and could even be mutually exclusive.
%
In FSL, we have two datasets during the training stage, i.e. a training set and a support set. 
The training set contains data from several source domains, $s_j,\,j\in [m]$ with $n_j$ annotated IID samples $(x_i,y_i) \stackrel{d}{\sim} p_{s_j}(x,y),\,\,i\in [n_j]$. 
The support set contains $k$-shot samples per class in the target domain, namely, $(x_i,y_i) \stackrel{d}{\sim} p_t(x,y),\,\,i \in [k \times |\mathcal{Y}^t|]$.
For testing, we have a test set with $n_t$ samples in the target domain.
Our goal is to learn a predictor $f(\cdot)$ so the expected loss $\bar{L}_t=\mathbb{E}_{(x,y)\sim p_t} \mathbbm{1}_{\{f(x)\neq y\}}$ is small. 
The $k$-shot samples in the support set are insufficient to learn a model for the new target space from scratch so the problem calls for techniques that can generalize from source datasets.

\noindent
\textbf{Zero-Shot Learning (ZSL)}: 
Again $p_s(x,y) \neq p_t(x,y)$ as in FSL. But in contrast to FSL we do not see annotated examples from target domain to help make a prediction. 

For training, a sub-collection, $\cY^s \subset \cY$ of so called seen classes are only available and no other data from unseen class, i.e., no input data associated with $\cY \setminus \cY^s$ are available. To help train predictors, semantic vectors, $\sigma_y \in \Sigma$ for $y \in \cY^s$ are provided and the semantic vectors and labels are in one-to-one correspondence. The source distribution is characterized as $p_s(x,\sigma) \propto p(x,\sigma)\mathbbm{1}_{\{\sigma_y: y \in \cY^s\}}$ and we obtain $n_s$ IID instances $(x_i,\sigma_{y_i})\sim p_s,\,\,i\in[n_s]$ for training. At test time we have full access to the semantic set $\Sigma$. In ZSL given an input instance $x \in \cX$ from target unseen set namely the associated label $y \in \cY^t = \cY \setminus \cY^s$, our goal is to train a predictor, $f(x)$ that minimizes expected loss: $\bar{L}_t=\mathbb{E}_{(x,\sigma)\sim p_t} \mathbbm{1}_{\{f(x)\neq \sigma\}}$, where $p_t \propto p(x,\sigma)\mathbbm{1}_{\{\sigma_y: y \not \in \cY^s\}}$.       

\textbf{Generalized Zero-Shot Learning (GZSL)}: While the training setup is identical to ZSL, at test-time, the input instances can be drawn from both seen and unseen object classes. Our goal is to minimize $\bar{L}_t=\mathbb{E}_{(x,\sigma)\sim p} \mathbbm{1}_{\{f(x)\neq \sigma\}}$. 

\subsection{Overview of Proposed Approach}
\label{sec:3.2}
The overall structure of our model is illustrated by {\it Figure~}\ref{fig:model}. 
The proposed model consists of a cascade of functions, including a part-feature extractor, a part-probability encoder, and a task specific predictor designed for different applications, e.g. GZSL, FSL and DA.

Specifically, let $\cD_{tr}$ denote the ordered pair of available input-output training instances, and $\cX_{tr}$ the corresponding input training instances.
For each input instance $x$, the part-feature extractor outputs $M$ attention regions and associated features, $z(x) = [z_m(x)]_{m \in [M]}, z_m(x) \in \mathbb{R}^{C}$, where the attention regions focus on different foreground object parts and have negligible overlap in the image space. 
For each part, the part-probability encoder aims to discover $K$ proto-typical atoms among the part-features in $\mathcal{X}_{tr}$, and project
each part feature vector $z_m$ on to such a dictionary of atoms $D_m$, resulting in a probability vector $\pi_m(x) \in \mathbb{R}^K, K \ll C$. 
The collection of part probability vectors $\pi(x)=[\pi_m(x)]_{m\in[M]}$ is then input into the task specific predictor $V(\pi(x))$,  which outputs a class label. 
%

The system is then trained to enforce three objectives: (1) the part-feature extractor should output diverse and discriminative attention regions
that focus on common object parts prevalent on most instances in $\cX_{tr}$; (2) the primitive proto-typical atoms should be representative to reconstruct the original part-features; and finally, (3) the predictor should be customized and optimized for each specific task.

\textbf{Prototypical Part Mixture Representation.} We build intuition into how our proposed scheme leads to good generalization on the proposed problems. As such, each atom in the dictionary can be viewed as a prototypical part-type. Specifically, we assume part-features $z_m$ in PTE can be represented by a Gaussian mixture of part-types. In other words, $z_m \sim \sum_k \pi_{k,m} {\cal N}(D_{k,m},\gamma^2 I)$, where $\pi_{k,m}$ represents the probability part $m$ belongs to component $k$ of the Gaussian component $D_{k,m}$ as shown in {\it Figure~}\ref{fig:embedding}.

\textbf{Conditional Independence.} From a probabilistic perspective we are placing a Markov chain structure on the relationship between input and output random variables (where, following convention, upper-case letters denote random variables):
$X \longleftrightarrow \pi(X) \longleftrightarrow Y$. Thus $p(y \mid \pi(x),x)=p(y \mid \pi(x))$ and so $\pi(x)$ serves as a sufficient statistic and the only uncertainty that remains is to identify the prediction map from $\pi(x)$ to the labels $y \in \cY$ at test-time.

\textbf{Discussion:} How is the mixture representation effective?\\
Suppose $p_s(x\mid y) \neq p_t(x\mid y)$, we could attempt to learn a mapping, $T(x)$ on the high-dimensional feature space so that for $x \sim p_t$ we have $T(x) \sim p_s(x)$, which we view as somewhat difficult. On the other hand, our proposed method learns and freezes LDVA encoding representing a composition of parts and offers benefits. 

\textbf{A.} \textit{Low-dimensionality.} The backbone network producing LDVA encoding is frozen at test-time. So learning a predictor on $\pi(x)$ requires relatively fewer examples. \\   \textbf{B.} \textit{Compositional Uniqueness.} Attention regions are sufficiently representative of important aspects of objects in terms of discriminability of objects \cite{zheng2017learning}. When the associated dictionary for each attention region are sufficiently descriptive, our visual encoding in terms of mixture composition uniquely describes different classes. \\
    \textbf{C.} \textit{Inter and Intra-Class Variances.} Intra-class variance arises from variance in visual appearance of a part-type within the same class and manifests in terms of the strength of the presence of the part-type in the input instance. On the other hand inter-class variance arises from the absence of parts or part-types, which results in smaller similarity in the visual encoding (see {\it Figure~\ref{fig:pi}}).

\textbf{Benefits of LDVA Encoding.}
In particular, under UDA we get $\pi(x)\approx \pi(x')$ for $x \sim p(\cdot | y),\, x'\sim p(\cdot | y)$ requiring no further alignment.
In FSL, we see new objects but as a consequence of (\textbf{B.}) these new objects are unique in terms of composition and furthermore as a result of (\textbf{C.}) are well separated. In ZSL we are given semantic vectors. Nevertheless, due to (\textbf{B.}) our representation closely mirror semantic vectors. Indeed, for many datasets, human-labeled semantic components are based on presence of visual parts in the class, and thus well-matched to LDVA encoding.

\subsection{Model and Loss Parameterization}
\noindent \textbf{Part-Feature Extractor}:  Inspired by \cite{zheng2017learning}, we use a multi-attention convolutional neural network (MA-CNN) to map input images into a finite set of part feature vectors, $z_m \in \mathbb{R}^C$. Specifically, it contains a global feature extractor $E$ and a channel grouping model $G$, where $E(x) \in \mathbb{R}^{W \times H \times C}$ is a global feature map, and $G(E(x)) \in \mathbb{R}^{M\times C}$ is a channel grouping weight matrix. We then calculate an attention map $A_m(x) \in \mathbb{R}^{W \times H}$ for the $m$-th part:
\begin{equation}
    A_m(x) = \mathrm{sigmoid}\big(\sum_c G_{m,c}(x) \times E_c(x)\big)
\end{equation}
The part feature $z_m \in \mathbb{R}^C$ is then calculated as:
\begin{equation}
z_{m,c} = \sum_{w,h} [A_m(x) \odot E_c(x)]_{(w, h)}  , \quad \forall c \in [C]
\end{equation}
where $\odot$ is the element-wise multiplication. We parameterized $E(\cdot)$ by the ResNet-34 backbone (to $conv5\_x$), and $G(\cdot)$ by a fully-connected layer.

To encourage a part-based representation $z_m$ to be learned, we follow \cite{zheng2017learning}.  Since $z_m$ can be decomposed into $A_m(x)\odot E(x)$, we want to force the learned attention maps $A_m$ to be both  compact within the same part, and divergent among different parts. We define $\ell_{part}$ to be:
\begin{equation}
    \ell_{part}(x) = \sum_m (L_{dis} (A_m (x)) + \lambda L_{div} (A_m (x))) \label{eq:loss_part}
\end{equation}
where the compact loss $L_{dis} (A_m)$ and divergent loss $L_{div} (A_m)$ are defined as ($x$ is dropped for simplicity):
\begin{align}
     L_{dis}({A_m}) &= \sum_{w,h} A_{m}^{w,h}[\|w - w^*\|^2 + \|h - h^*\|^2] \label{eq:loss_Dis} \\
     L_{div}({A_m}) &= \sum_{w,h} A_{m}^{w,h}[\max_{n, n\ne m}A_n^{w,h} - \zeta] \label{eq:loss_Div}
\end{align}
where $A_{m}^{w,h}$ is the amplitude of $A_m$ at coordinate $(w,h)$, and $(w^*, h^*)$ is the coordinate of the peak value of $A_m$, $\zeta$ is a small margin to ensure the training robustness.

\noindent \textbf{Part-Probability Encoder}:  Our Gaussian assumption leads us to an auto-encoder implementation to map the high-dimensional part-feature $z_m(x) \in \mathbb{R}^C$ into the low-dimensional probability $\pi_m(x) \in \mathbb{R}^K$.
Specifically, for part $m$, given the part features $z_m(x) \in \mathbb{R}^{C}$, we define a projection matrix $P_m \in \mathbb{R}^{K \times C}$, such that:
\begin{equation}
\setlength{\abovedisplayskip}{5pt}
\setlength{\belowdisplayskip}{3pt}
    \pi_m(x) = P_m z_m(x)
    \label{eq:pi}
\end{equation}

{\it Guassian Mixture Condition}: Our Gaussian mixture assumption (Sec.\ref{sec:3.2}) implies the following condition should hold:
\begin{equation}
    z_m(x) \approx D_m^\top \pi_m(x) ,
    \label{eq:z}
\end{equation}
where $D_m \in \mathbb{R}^{K \times C}$ is a 'fat' matrix ($K \ll C$) of Gaussian components for part $m$, i.e. $D_m = [D_{k,m}]_{k\in[K]}$. 

Viewing $P_m$ and $D_m$ as model parameters, our training objective can be naturally written in the form of an auto-encoder, where $P_m$ is the encoder and $D_m$ is the decoder:
\begin{equation}
\resizebox{.5 \textwidth}{!} 
{$\ell_{prob}(x) = \sum_m \Big( \| z_m(x) - D_m^\top P_m z_m(x) \|^2  + \lambda \| P_m \|^2 + \lambda \| D_m \|^2 \Big).$}
   \label{eq:prob}
\end{equation}

\noindent \textbf{Task Specific Predictors}: The part-probability $\pi$ serves as an input to a task specific predictor $V(\pi)$.

\noindent {\it Generalized Zero-Shot Learning}: For GZSL, $V(\pi)$ is a semantic prediction model parameterized by a neural network to project $\pi$ into the semantic space ${\Sigma}$.  Given an input image $x$ and its semantic attribute $\sigma_y$, the loss for training the GZSL predictor, with $\eta$ as margin parameter is modeled as:
\begin{align}
\resizebox{.5 \textwidth}{!} 
{$
\ell_{GZSL}(x,y) = \sum_{y' \in \mathcal{Y}} [\eta \mathbbm{1}[y'=y] +\sigma_{y}^\top V(\pi(x)) 
-\sigma_y^\top V(\pi(x))]_+ $}
 \label{eq:pi_to_s}
 \vspace{-0.1in}
\end{align}

\noindent {\it Few-Shot Learning}: For FSL, we have different implementations for the predictor in the source domain and the target domain. For an input-output pair $(x,y)$ in the source domain training set, $V(\pi)$ is a classification model parameterized by a neural network to project $\pi$ into the class label space $\mathcal{Y}^s$. The training loss is simply a cross-entropy loss:
\begin{equation}
\ell_{FSL}(x,y) = \text{CE}(V(\pi(x)), o(y)), 
\end{equation}
where  $\text{CE}(\cdot,\cdot)$ is the cross-entropy, $o(\cdot)$ is the one-hot encoding function.
After training, we calculate the average $\bar{\pi}_y$ representation for $K$-shot samples in the target domain support set, and build a nearest neighbour classifier for testing, i.e.
$V(\pi(x)) = \arg\min_{y\in\mathcal{Y}^t} | \pi(x) - \bar{\pi}_y|^2$.

\noindent {\it Domain Adaptation}: For DA, $V(\pi)$ is a classification model parameterized by a neural network to project $\pi$ into the class label space $\mathcal{Y}$. In DA we have training samples from both source domain $\mathcal{D}_s$ and target domain $\mathcal{D}_t$, where $D_t$ has no class label available during training. Inspired by \cite{chadha2018improving,saito2017asymmetric}, we estimate psuedo-labels for the target domain samples with the current classification model $V(\pi)$ and further optimizes the following loss:
\begin{align}
\ell_{DA}(x,y) = & \mathbbm{1}[x \in \mathcal{D}_s]\text{CE}(V(\pi(x)), o(y)) \nonumber \\
              &+ \mathbbm{1}[x \in \mathcal{D}_t]\text{CE}(V(\pi(x)), o(\hat{y})), 
              \label{eq:da}
\end{align}
where $\hat{y} = \arg\max_y V(\pi(x))_y$, and $V(\pi(x))_y$ is the $y$-th element in the $V(\pi(x))$ vector. 

By pseudo-labelling target samples, we aim to align the class level source-target distributions, i.e. aligning $p_s(x|y)$ and $p_t(x|y)$, and meanwhile minimize the entropy of the prediction distributions, such that a discriminative $\pi$ representation that convey confident decision rules can be learnt.

\noindent \textbf{End-to-End Training}: We train our system discriminatively by employing three loss functions.  In particular, suppose the part-feature extractor is parameterized by $\Theta$, 
the part-probability encoder by $([D_m, P_m])$, the predictor by $\alpha$,  the overall training objective is:
{
\setlength{\abovedisplayskip}{5pt}
\setlength{\belowdisplayskip}{3pt}
\begin{align}
\min_{\theta,\alpha, [D_m, P_m]}\sum_{(x,y) \in \cD_{tr}}& \ell_{part}(x; \Theta) +  \ell_{prob}(x; [D_m, P_m], \Theta) \nonumber \\  &+ \ell_{task}(x,y; \alpha, [D_m, P_m], \Theta)
\end{align}
where $\ell_{task} \in \{\ell_{GZSL}, \ell_{FSL}, \ell_{DA}\}$.
}

\subsection{Implementation Details}
We set the number of parts $M$ to 4 and in each part the number of prototypes $K$ is set to 16. $\zeta$ in Eq.(\ref{eq:loss_Div}) is empirically set to 0.02. For FSL, we set the input size to be [224 $\times$ 224], and $\lambda$ in Eq.(\ref{eq:loss_part}) is 2; 
for GZSL, our model takes input image size as [448 $\times$ 448] and $\lambda$ is set to 5;
For DA, the input image size is [224 $\times$ 224] and $\lambda$ is set to 2. The task-specific predictor $V(\cdot)$ for both GZSL and DA is implemented by a two FC-layer neural network with ReLU activation, the number of neurons in the hidden layer is set to 32.

Our model takes an alternative optimization approach to minimize the overall loss. In each epoch, we update the weights in two steps. In step.A, only the weights of channel grouping $G(\cdot)$ is updated by minimizing $\ell_{part}$. In step.B, we freeze the weights of $G(\cdot)$ and update all the other modules. Adam optimizer is used in each step.

\begin{figure}
    \centering
    \includegraphics[width=\linewidth]{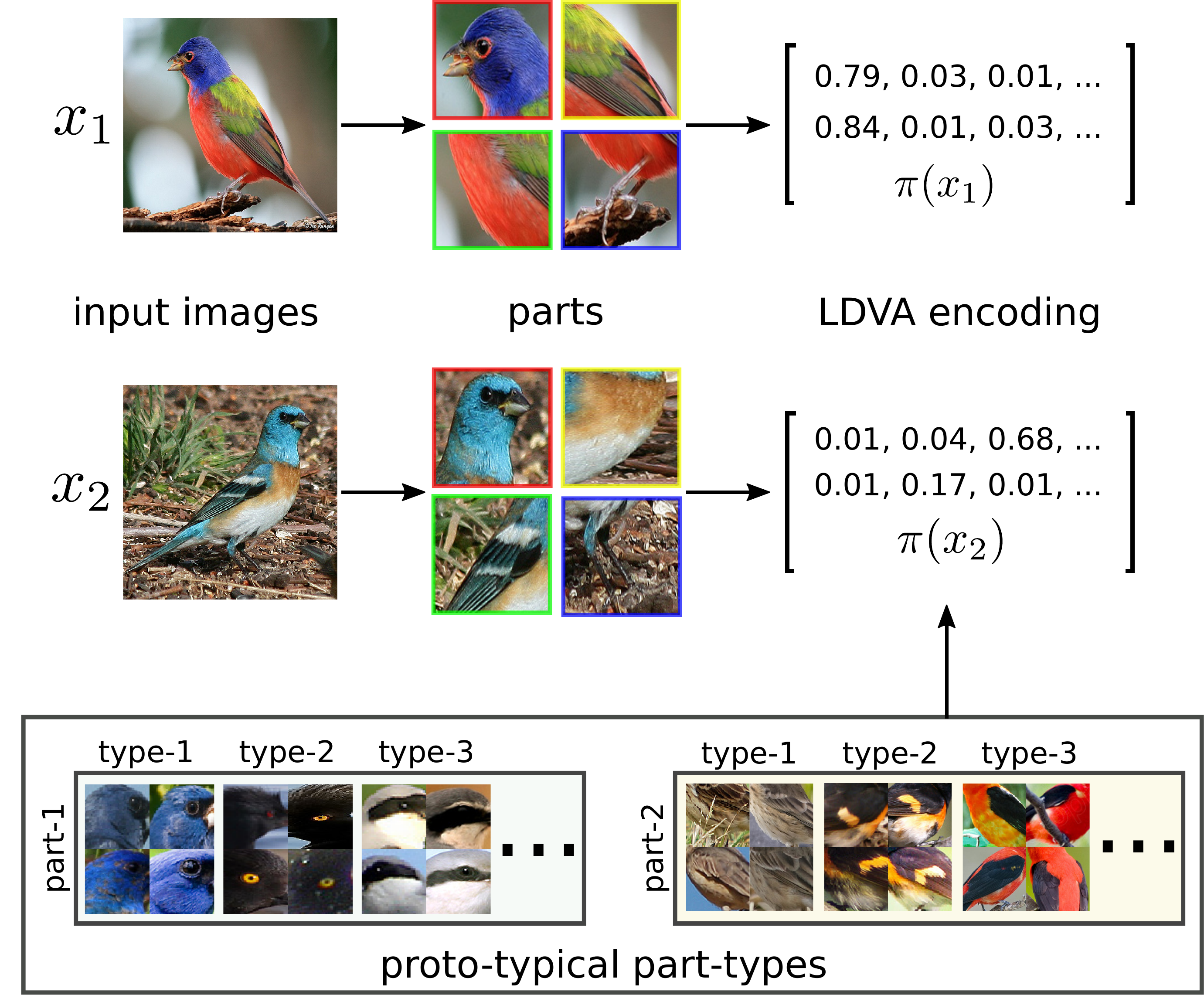}
    \vspace{-0.5cm}
    \caption{LDVA is generated so that the mixture of the proto-typical part types represents the corresponding part. The objects have similar representation if they have similar visual parts. The resulting LDVA encoding also has a smaller gap to the semantic attributes in the GZSL setting, e.g. beak-color, wing-color, etc, compared to the original high-dimensional features.}
    \label{fig:embedding}
\end{figure}

\vspace{-0.2cm}
\section{Experiments}
\label{experiments}
\subsection{Few-Shot Learning}
{\noindent \bf Datasets.} We first evaluate the few shot learning performance of the proposed model on two benchmark datasets: Omniglot\cite{lake2015human} and {\it mini}ImageNet\cite{vinyals2016matching}. Omniglot consists of 1623 characters from 50 alphabets. Each character (class) contains 20 handwritten images from people. {\it mini}ImageNet is a subset of ImageNet\cite{ILSVRCarxiv14} which contains 60,000 images from 100 categories.

{\noindent \bf Setup.} We follow the same protocol in \cite{sung2018learning}. For Omniglot, the dataset is augmented with new classes through 90\degree, 180\degree\ and 270\degree \ rotations of existing characters. 1200 original classes plus rotations are selected as training set and the remaining 423 classes with rotations are test set. For {\it mini}ImageNet, the dataset is split into 64 training, 16 validation and 20 testing classes. The model will only be trained on training set and the validation set is for examining the training performance.

We evaluate the 5-way accuracy on {\it mini}ImageNet and 5-way plus 20-way accuracy on Omniglot. 1-shot and 5-shot learning performance is evaluated in each setting. For {\it m-way k-shot} learning, in each test episode, {\it m} classes will be randomly selected from the test set, then {\it k} samples will be drawn from these classes as support examples, and 15 examples will be drawn from the rest images to construct the test set. We run 1000 and 600 test episodes on Omniglot and {\it mini}ImageNet, respectvely, to compute the average classification accuracy.

{\noindent \bf Training Details.} Our model is trained for 80 and 30 epochs on Omniglot and {\it mini}ImageNet, repectively. The learning rate for step.A is set to 1e-6, and the learning rate of step.B is 1e-4 for Omniglot and 1e-5 for {\it mini}ImageNet. On {\it mini}ImageNet, the weights for the feature extractor $E(\cdot)$ is pretrained on the training split.

{\noindent \bf Competing Models.} We list here the state-of-the-art methods we compare to:
Matching Nets\cite{vinyals2016matching}, Prototypical Nets\cite{snell2017prototypical}, Meta Nets\cite{munkhdalai2017meta},
MAML\cite{finn2017model}, Relation Nets\cite{sung2018learning}, TADAM\cite{oreshkin2018tadam}, LEO\cite{rusu2018meta}, and EA-FSL\cite{ye2018learning}. Their description can be found in Sec.\ref{related}.

\begin{table}[t]
    \centering
    \small
    \renewcommand{\arraystretch}{1.3}
    \setlength{\tabcolsep}{0.1cm}
    \begin{tabular}{|l|c c|c c|c c|}
        \hline
        \multirow{3}{*}{\bf Methods} & \multicolumn{4}{c|}{Omniglot} & \multicolumn{2}{c|}{{\it mini}ImageNet} \\
        \cline{2-7}
        & \multicolumn{2}{c|}{5-way Acc.} & \multicolumn{2}{c|}{20-way Acc.} & \multicolumn{2}{c|}{5-way Acc.} \\
         & 1-shot & 5-shot & 1-shot & 5-shot & 1-shot & 5-shot \\
        \hline
        Matching Nets & 98.1 & 98.9 & 93.8 & 98.5 & 43.6 & 55.3 \\
        Meta Nets & 99.0 & - & 97.0 & - & 49.2 & - \\
        MAML & 98.7 & \bf 99.9 & 95.8 & 98.9 & 48.7 & 63.1 \\
        Prototypical Nets & 98.8 & 99.7 & 96.0 & 98.9 & 49.4 & 68.2 \\
        Relation Nets & \bf 99.6 & 99.8 & \bf 97.6 & 99.1 & 50.4 & 65.3 \\
        TADAM & - & - & - & - & 58.5 & 76.7 \\
        LEO & - & - & - & - & 61.7 & 77.6 \\
        EA-FSL & - & - & - & - & \bf 62.6 & 78.4 \\
        \hline
        Ours & 98.9 & 99.8 & 96.5 & \bf 99.3 & 61.7 & \bf 78.7 \\
        \hline
    \end{tabular}
    \caption{FSL classification results on Omniglot and {\it mini}ImageNet. \label{tab:fsl}}
\end{table}

\begin{table*}[t]
    \centering
\renewcommand{\arraystretch}{1}
\setlength{\tabcolsep}{0.3cm}
\scalebox{1.0}{
    \begin{tabular}{|l| c c c|c c c|c c c|}
        \hline
        \multirow{2}{*}{\bf{Methods}} & \multicolumn{3}{c|}{\bf{CUB}} & \multicolumn{3}{c|}{\bf{AWA2}} & \multicolumn{3}{c|}{\bf{aPY}} \\
         & ts & tr & H & ts & tr & H & ts & tr & H\\
        \hline
        SJE\cite{akata2015evaluation} & 23.5 & 59.2 & 33.6 & 8.0 & 73.9 & 14.4 & 3.7 & 55.7 & 6.9 \\
        SAE\cite{kodirov2017semantic} & 7.8 & 54.0 & 13.6 & 1.1 & 82.2 & 2.2 & 0.4 &   80.9 & 0.9 \\
        SSE\cite{zhang2015zero} & 8.5 & 46.9 & 14.4 & 8.1 & 82.5 & 14.8 & 0.2 & 78.9 & 0.4 \\
        GFZSL\cite{verma2017simple} & 0.0 & 45.7 & 0.0 & 2.5 & 80.1 & 4.8 & 0.0 & {\bf\color{blue}83.3} & 0.0 \\
        CONSE\cite{norouzi2013zero} & 1.6 & 72.2 & 3.1 & 0.5 & 90.6 & 1.0 & 0.0 & {\bf\color{red}91.2} & 0.0 \\
        ALE\cite{akata2016label} & 23.7 & 62.8 & 34.4 & 14.0 & 81.8 & 23.9 & 4.6 & 73.7 & 8.7 \\
        SYNC\cite{changpinyo2016synthesized} & 11.5 & 70.9 & 19.8 & 10.0 & 90.5 & 18.0 & 7.4 & 66.3 & 13.3 \\
        DEVISE\cite{frome2013devise} & 23.8 & 53.0 & 32.8 & 17.1 & 74.7 & 27.8 & 4.9 & 76.9 & 9.2 \\
        PSRZSL\cite{Annadani_2018_CVPR} & 24.6 & 54.3 & 33.9 & 20.7 & 73.8 & 32.3 & 13.5 & 51.4 & 21.4 \\
        SP-AEN\cite{chen2018zero} & 34.7 & 70.6 & 46.6 & 23.3 & 90.9 & 37.1 & 13.7 & 63.4 & 22.6 \\
        \hline
        {\it Generative ZSL} & & & & & & & & & \\
        GDAN\cite{huang2018generative} & 39.3 & 66.7 & 49.5 & 32.1 & 67.5 & 43.5 & \bf {\color{blue}30.4} & 75.0 & \bf {\color{blue}43.4} \\
        CADA-VAE\cite{schonfeld2018generalized} & 51.6 & 53.5 & 52.4 & 55.8 & 75.0 & 63.9 & - & - & - \\
        3ME\cite{felix2019multi} & 49.6 & 60.1 & 54.3 & - & - & - & - & - & - \\
        SE-GZSL\cite{Verma_2018_CVPR} & 41.5 & 53.3 & 46.7 & \bf \color{blue}{58.3} & 68.1 & 62.8 & - & - & - \\
        LSD\cite{dong2018learning} & 53.1 & 59.4 & 56.1 & - & - & - & 22.4 & 81.3 & 35.1 \\
        DA-GZSL\cite{atzmon2018domain} & 47.9 & 56.9 & 51.8 & - & - & - & - & - & - \\
        \hline
        {\it Trans-ZSL} & & & & & & & & & \\
        DIPL\cite{NIPS2018_7380} & 41.7 & 44.8 & 43.2 & - & - & - & - & - & - \\
        TEDE\cite{zhang2018effective} & \bf \color{blue}{54.0} & 62.9 & \bf \color{blue}{58.1} & \bf\color{red}{68.4} & \bf\color{red}{93.2} & \bf\color{red}{78.9} & 29.8 & 79.4 & 43.3 \\
        \hline
        Ours & 33.4 & {\bf\color{red}87.5} & 48.4 & 41.6 & {\bf\color{blue}91.3} & 57.2 & 24.5 & 72.0 & 36.6 \\
        Ours + CS & {\bf\color{red}{59.2}} & {\bf\color{blue}74.6} & {\bf\color{red}66.0} & 54.6 & 87.7 & {\bf\color{blue}67.3} & {\bf\color{red}41.1} & 68.0 & {\bf\color{red}51.2} \\
        \hline
    \end{tabular}}
    \caption{GZSL results on CUB, AWA2 and aPY. ts = test classes (unseen classes), tr = train classes (seen classes), H = harmonical mean. The accuracy is class-average Top-1 in \%. The highest accuracy is in \textcolor{red}{red} color and the second is in \textcolor{blue}{blue} (better viewed in color).}
    \label{tab:gzsl}
\end{table*}

{\noindent \bf Results.} Few shot learning results are shown in {\it Table~\ref{tab:fsl}}. On both datasets, our model reaches the same level accuracy as other state-of-the-art methods. Specifically, on {\it mini}ImageNet, our model obtain 78.7\% for 5-shot learning scanerio, which supass the second best model with an absolutely margin 0.3\%. On omniglot, the accuracy for 20-way 5-shot learning is improved to 99.3\%.

Compared with other methods which process the high-dimensional visual features or utilize meta-learning strategy, our model leverages the LDVA representations to reduce the inter-class variance for novel categories. That is, because of the unique composition for each class, the distance between examples in the same class is smaller than the high-dim features. This results in the good performance in LDVA even only a simple nearest neighbor classifier is applied. In addition, since a universal part prototypes is learned from the seen classes, our model does not require any meta-training or fine-tune on the unseen categories, while meat-learning based methods need to dynamically adapt their model based on the feedback from new tasks.

\subsection{Generalized Zero-Shot Learning}
{\noindent \bf Datasets.}
The performance of our model for GZSL is evaluated on three commonly used benchmark datasets: {\it Caltech-UCSD Birds-200-2011} (CUB) \cite{WahCUB_200_2011}, {\it Animals with Attributes 2} (AWA2) \cite{xian2018zero} and Attribute Pascal and Yahoo (aPY) \cite{farhadi2009describing}. CUB is a fine-grained dataset consisting of 11,788 images from 200 different types of birds. 312-dim semantic attributes are annotated for each category. AWA2 is a coarse-grained dataset which has 37,222 images from 50 different animals and 85-dim class-level semantic attributes. aPY contains 20 Pascal classes and 12 Yahoo classes. It has 15,339 images in total and 64-dim semantic attributes are provided.

{\noindent \bf Setup.}
Recent works~\cite{xian2018zero} have shown that the conventional ZSL setting is overly optimistic because it leverages absence of seen classes at test-time and there is an emerging consensus that methods should focus on the generalized ZSL setting.  
We thus evaluated under the GZSL setting.
Following the protocol in \cite{xian2018zero}, we evaluate the average-class Top-1 accuracy on unseen classes (ts), seen classes (tr) and the harmonic mean (H) of ts and tr.

{\color{black} It has been observed the scores for seen classes are often greater than unseen in GZSL methods \cite{Chao2016AnES}, which results in poor performance. Calibrated Stacking(CS) is proposed in \cite{Chao2016AnES} to balance the performance between seen and unseen classes by calibrating the scores of seen classes. As tabulated in {\it Table~\ref{tab:gzsl}}, in addition to our original model, we also apply CS into our model to alleviate this imbalance, denoted as Ours+CS. The parameters for CS is chosen via cross validation.}

{\noindent \bf Training Details.}
Our models are trained for 120, 100 and 110 epochs on CUB, AWA2 and aPY, respectively. The learning rate for step.A and step.B is set to 1e-6 and 1e-5.

{\noindent \bf Competing Models.}
We compare against state-of-the-art approaches. Comparisons are not all one-to-one since some of these approaches utilize different assumptions: {\color{black}(1) learns a compatibility function between the visual and semantic representations: SJE\cite{akata2015evaluation}, ALE\cite{akata2016label},  DEVISE\cite{frome2013devise}, SAE\cite{kodirov2017semantic}, SSE\cite{zhang2015zero}, CONSE\cite{norouzi2013zero}, SYNC\cite{changpinyo2016synthesized}, GFZSL\cite{verma2017simple}, PSRZSL\cite{Annadani_2018_CVPR}, and SP-AEN\cite{chen2018zero}. {\bf Our method also uses this strategy.} (2) Generative model based methods ({\it Generative-ZSL}). These methods synthesize unseen examples or features using generative models like GAN and VAE thus require unseen class semantics in the training time: GDAN\cite{huang2018generative}, CADA-VAE\cite{schonfeld2018generalized}, 3ME\cite{felix2019multi}, SE-GZSL\cite{Verma_2018_CVPR}, LSD\cite{dong2018learning}, and DA-GZSL\cite{atzmon2018domain}.
(3) transductive ZSL ({\it Trans-ZSL}). These methods work in a transuctive setting in which even examples for unseen classes are available during training: DIPL\cite{NIPS2018_7380}, and TEDE\cite{zhang2018effective}.}

{\noindent \bf Results.}
{\color{black} Results for GZSL are in {\it Table~\ref{tab:gzsl}}. Without the calibrated stacking, our model (ours) reaches 48.4\% on CUB, 57.2\% on AWA2 and 36.6\% on aPY for the harmonic mean (H), which outperforms all other compatibility function based methods. After the scores are calibrated, our model (ours+CS) obtains 66.0\%, 67.3\% and 51.2\% for the harmonic mean, respectively, which outperforms all other competing models except TEDE on AWA2. Specifically, on CUB, Ours+CS surppasses the 2-nd best result by a margin of 5.2\% on ts, 11.7\% on tr, and 7.9\% on H. On aPY, our models improves the accuracy for unseen classes from 30.4\% to 41.1\%, resulting in a 7.8\% increase on harmonic mean.

It is worth noting that, the {\it Generative ZSL} and {\it Trans-ZSL} methods always obtain higher accuracy than compatibility function methods, except for our models. This is because the generative and trans-ZSL methods have access to additional information of unseen classes during training. However, this assumption is too optimistic in real world ZSL scenario since it is unlikely to have full knowledge of all unseen categories in the training stage. In contrast, our models can be applied in the scenario where novel classes may only appear at test time. Still, by only leveraging seen classes knowledge, our models obtain competitive and even better performance than generative and trans-ZSL methods.}

The success of our model can be attributed primarily to the proposed LDVA representation, which resembles the components of the semantic attributes. For example, in {\it Figure~\ref{fig:embedding}}, we visualize the part attentions discovered by our model and several semantic attributes for the class {\it `Painted\_Bunting'} in CUB dataset. Our model learns the part areas around ``head'', ``wing'', ``body'', and ``feet'', based on which are most semantic attributes annotated (e.g. crow color: blue, wing color: green, etc.). Via the prototype encoding, our visual attributes mirror the representation of semantic vectors, thus mitigating the large gap between the semantic attributes and high-dimensional visual features.

\begin{table}[t]
    \centering
    \renewcommand{\arraystretch}{1.16}
    \setlength{\tabcolsep}{0.15cm}
    \begin{tabular}{|l| c c c|}
        \hline
        \bf Methods & M $\rightarrow$ U & U $\rightarrow$ M & S $\rightarrow$ M \\
        \hline
        Gradient reversal & 77.1 & 73.0 & 73.9\\
        Domain confusion & 79.1 & 66.5 & 68.1 \\ 
        CoGAN & 91.2 & 89.1 & - \\
        ADDA  & 89.4 & 90.1 & 76.0 \\
        DTN & - & - & 84.4 \\
        UNIT & 96.0 & 93.6 &  90.5 \\
        CyCADA & 95.6 & 96.5 & 90.4 \\
        MSTN & 92.9 & - & 91.7 \\
        Self-ensembling & \color{blue}{\bf 98.3} & \bf{\color{red}99.5} & \bf{\color{red}99.3}\\
        \hline
        Ours (source $\pi$) & 94.8 &  96.1 & 82.4\\
        Ours (joint $\pi$) & \color{red}{\bf{98.8}} & \bf{\color{blue}96.8} & \bf{\color{blue}95.2} \\
        \hline
    \end{tabular}
\caption[caption]{DA classification results. M = MNIST, U = USPS, S = SVHN. The highest accuracy is in \textcolor{red}{red} color and the second is in \textcolor{blue}{blue} (better viewed in color). Self-ensembling, unlike other methods, leverages data-augmentation and reports accuracy numbers that are evidently higher than those obtained in the fully supervised case for $U\rightarrow M,\,S \rightarrow M$. \label{tab:da}}
\end{table}

\begin{figure}[t]
\centering
\includegraphics[width=0.48\textwidth]{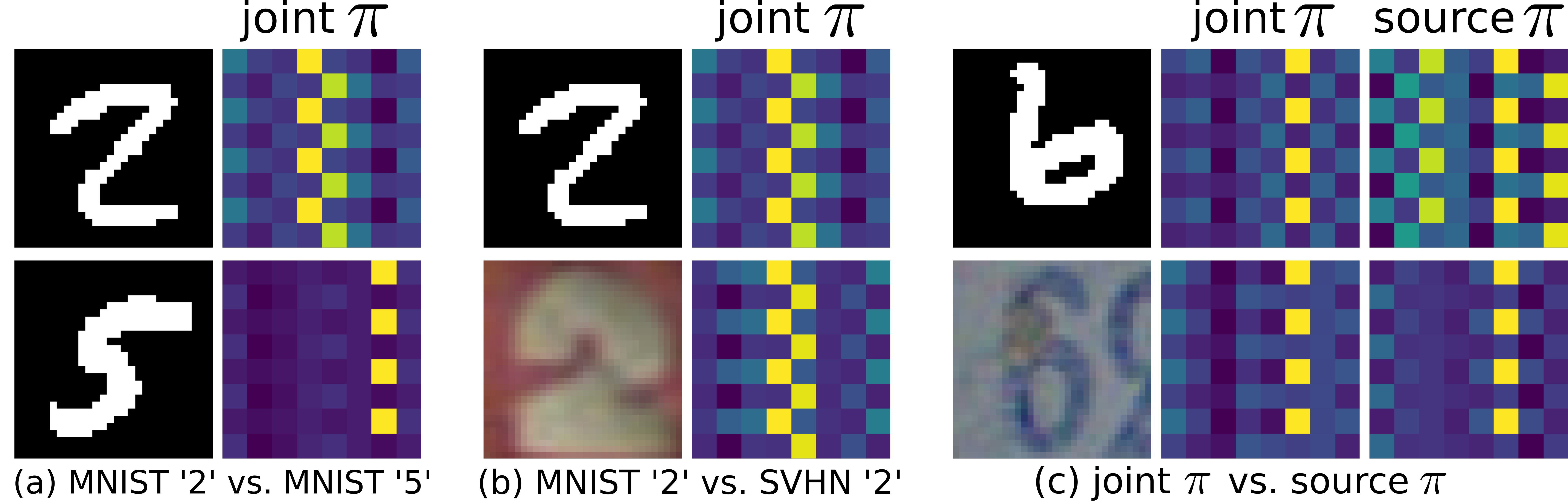}
\vspace{-0.4cm}
\caption{Proposed LDVA encoding $\pi$ on digit datasets. The $64$-dimensional $\pi$ vector is reshaped to a $8\times8$ matrix for better visualization. For all three examples (a-c), $\pi$ is trained for SVHN$\to$MNIST experiment. }
\label{fig:pi}
\end{figure}

\subsection{Domain Adaptation}
{\noindent \bf Datasets.}
We evaluate our proposed model in unsupervised domain adaptation task between three digits datasets: MNIST\cite{lecun1998gradient}, USPS and SVHN\cite{netzer2011reading}. Each dataset contains 10 classes of digit numbers (0-9). MNIST and USPS are handwritten digits while SVHN is obtained from house number in google street view images.  

{\noindent \bf Setup.}
We follow the same protocol in \cite{tzeng2017adversarial}, where the adaptation in three directions are validated: MNIST$\to$USPS, USPS$\to$MNIST, and SVHN$\to$MNIST. In the experiments, two variants of our model are evaluated: (1) {\bf Ours(source $\pi$)}: During training, the model is purely learned from source data. In this case, $\ell_{DA}$ reduces to a standard cross-entropy loss on the source domain. In test time, LDVA encoding for target data is based on the source visual encoder $P_m^s$. This model does not utilize any information from the unlabeled target data in the training. (2) {\bf Ours(joint $\pi$)}: This model learns the visual encoder from the joint dataset $\mathcal{D}_s\cup\mathcal{D}_t$ as described by  Eq.(\ref{eq:da}). 

{\noindent \bf Training Details.}
Ours (source $\pi$) is trained on the source domain dataset, as described above. The learning rate for step.A and step.B is 1e-6 and 1e-5. The training epochs are set to be 40, 20, and 40 on MNIST, USPS, and SVHN, respectively. For joint $\pi$, we first initialize with weights from trained on our source-$\pi$ model. Next, the model is trained on the joint dataset $\mathcal{D}_s\cup\mathcal{D}_t$ for 10 epochs. The learning rate for step.B is modified to 1e-6.

{\noindent \bf Competing Models.}
We compare against several state-of-the-art UDA methods:
Gradient reversal~\cite{ganin2016domain}, ADDA~\cite{tzeng2017adversarial},  Domain confusion~\cite{tzeng2015simultaneous},
CoGAN~\cite{liu2016coupled}, DTN~\cite{taigman2016unsupervised}, UNIT~\cite{liu2017unsupervised}, CyCADA~\cite{hoffman2017cycada}, MSTN~\cite{xie2018learning}, and Self-ensembling\cite{french2017self}.

{\noindent \bf Results.}
The results for DA are shown in {\it Table~\ref{tab:da}}. Specifically, Ours (source $\pi$) and Ours (joint $\pi$) reaches 94.8\%, 98.8\% on M$\to$U, 96.1\%, 96.8\% on U$\to$M, and 82.4\%, 95.2\% on S$\to$M. On M$\to$U, our method with jointly learned $\pi$ outperforms all other competing methods. On U$\to$M and S$\to$M, our method obtains the second best performance, only left behind Self-ensembling.
{\color{black}It is worth noting that Self-ensembling is a data augmentation technique which models the distortion in target data. Evidently, this technique for the specific dataset is so powerful that the reported accuracies are higher than those reported for a fully supervised model on target data. In contrast, our method learns a static universal representation for both source and target domain, which does not require the prior knowledge on the domain distortion. The data augmentation is complementary to our model and it can be expected that our model can also benefit from the increasing training data.}

The results demonstrate the benefits of proposed LDVA representation. Specifically, in the same domain, the LDVA representations for different classes are large enough to learn a good classifier. Meanwhile, 
the representations of the same class from different domains are much more similar than the high-dimensional features, resulting in a similar distribution for $\pi_s$ and $\pi_t$. The classifier on source domain are thus able to be applied on target domain. We also illustrate this effect in {\it Figure~}\ref{fig:pi}(a-b). As we can see the part probability vector of digit '2' in MNIST is very similar to SVHN, while quite different against the digit '5' in MNIST.

{\noindent \it Source vs. Joint $\pi$}. Our model with jointly learned $\pi$ outperforms purely source $\pi$ with 4.4\%, 0.7\% and 12.8\% absolute improvement on the three adaptation directions. This comparison shows that the cross-entropy loss using pseudo-label on the target domain in Eq.(\ref{eq:da}) helps the model learn a more universal prototypes, and hence reduces the distance between the representations of the same class. The model will benefit more when the domain shift is severe. As shown in {\it Figure~}\ref{fig:pi}(c), the part probability vector of digit '6' in MNIST is more similar to the one in SVHN in the joint $\pi$ space. This also results in the largest performance gap on S$\to$M, since SVHN is obtained from street view while MNIST and USPS are both handwritten digits. 

{\noindent \it Tolerance to Visual Distortions}.
Note that all of the competing methods are trained jointly on both source and target domains 
and so, comparing against our source-$\pi$ method is an unfair comparison. Still, what we see here from the first two experiments is that access to unlabeled target data is somewhat unnecessary if we adopt LDVA encoding. This points to the fact that mixture compositions are tolerant to visual distortions, which can be an issue for methods relying on transferring information in high-dimensions. On the other hand for the last experiment, the variance is significant and unannotated target data is useful.

\section{Conclusion}
\label{conclusion}
We proposed a novel method for computer vision problems, where new tasks and environments arise. In these cases, due to limited supervision on the target, training ``models from scratch,'' is impossible and methods that adapt existing models, trained on the presented training environment, to the new scenario are required. We propose a novel low-dimension visual attribute (LDVA) encoding method that represents the mixture composition of prototypical parts of any instance. The LDVA encodings are low-dimensional, are capable of uniquely representing new objects and are tolerant to visual distortions. We train an end-to-end model for a variety of tasks including domain adaptation, few shot learning and zero-shot learning. 
Our method outperforms state-of-art methods even though those methods are customized to the specific problem contexts (ZSL, FSL, DA).
%


\bibliography{related}
\bibliographystyle{icml2019}

\end{document}